%% file: main.tex
\documentclass[letterpaper,10pt,twocolumn,conference]{article} 
\usepackage{simpleConference}
\usepackage{times}
\usepackage{graphicx}
\usepackage{amssymb}
\usepackage{url,hyperref}
\usepackage{graphicx}
\usepackage{subcaption}
\usepackage{amsmath}
\usepackage{float}
\usepackage{xcolor}
\usepackage{multirow}
\usepackage{booktabs}
\usepackage{bm}
\usepackage[colorinlistoftodos]{todonotes}
\usepackage{cleveref}
\crefformat{section}{\S#2#1#3} 
\crefformat{subsection}{\S#2#1#3}
\crefformat{subsubsection}{\S#2#1#3}

\usepackage{cite}
\usepackage[colorinlistoftodos]{todonotes}

\graphicspath{ {images/} }

\newcommand{\nostarnote}[1]{}
\newcommand{\baad}[1]{} 
\usepackage{color}
\definecolor{Junaed_color}{RGB}{255,0,0} 

\newcommand{\eg}{\emph{e.g.}, }
\newcommand{\ie}{\emph{i.e.}, } 

\usepackage{soul}

\title{\LARGE \bf Real-Time Multi-Diver Tracking and Re-identification \\for Underwater Human-Robot Collaboration}

\author{Karin de Langis$^{1}$ and Junaed Sattar$^{2}$
\thanks{The authors are with the Department of Computer Science and Engineering, Minnesota Robotics Institute, University of Minnesota Twin Cities, Minneapolis, MN, USA.
{\tt\small \{$^{1}$dento019, $^{2}$junaed\} at umn.edu.}}
}

\begin{document}

\maketitle
\thispagestyle{empty}
\pagestyle{empty}

\input{tex/abstract}
\input{tex/intro}
\input{tex/related}
\input{tex/methodology}
\input{tex/experiments}

\input{tex/conclusion}

\newpage
\bibliographystyle{ieeetr}
\bibliography{citations}

\end{document}

%% file: tex/abstract.tex
\begin{abstract}
Autonomous underwater robots working with teams of human divers may need to distinguish between different divers, \eg to recognize a lead diver or to follow a specific team member. This paper describes a technique that enables autonomous underwater robots to track divers in real time as well as to reidentify them. The approach is an extension of Simple Online Realtime Tracking (SORT) with an appearance metric (deep SORT). Initial diver detection is performed with a custom CNN designed for realtime diver detection, and appearance features are subsequently extracted for each detected diver. Next, realtime tracking-by-detection is performed with an extension of the deep SORT algorithm. We evaluate this technique on a series of videos of divers performing human-robot collaborative tasks and show that our methods result in more divers being accurately identified during tracking. We also discuss the practical considerations of applying multi-person tracking to on-board autonomous robot operations, and we consider how failure cases can be addressed during on-board tracking.

\end{abstract}

%% file: tex/intro.tex
\section{INTRODUCTION}

The state of the art in multi-person visual tracking has greatly improved in both speed and accuracy in recent years~\cite{milan_mot16:_2016, luo_multiple_2014, leal2017tracking}. These improvements make multi-person trackers viable for use on realtime robotic platforms. However, utilizing multi-person tracking algorithms onboard autonomous robots, particularly in adverse conditions, is still an under-explored area \cite{islam_person_2018}. In this paper, we propose a realtime multi-person tracker suitable for autonomous underwater robots. 

This work was motivated by the need for underwater robots to distinguish between different human `teammates' in order to improve underwater human-robot collaboration. Underwater robots are utilized for a wide range of tasks, including data collection, ecological mapping, and wreck investigations (\eg \cite{hoegh2007coral,shkurti2012multi,bingham2010robotic}). These tasks frequently require collaboration between robots and human divers. When the robot collaborates with a team of divers, it is highly useful for the robot to be able to identify different divers: for example, the robot may need to follow a specific diver or to recognize a lead diver from whom to take instructions. 

In order to distinguish between different divers, the robot must continually detect divers who are present in images received from its cameras. The robot must also have a method to ``identify" detections; that is, the robot must keep track of each person it has seen and determine whether a detection corresponds to one of those people. This is roughly equivalent to the multi-object tracking (MOT) problem, which has been extensively studied in computer vision.

\begin{figure}
    \centering
    \includegraphics[width=0.8\linewidth]{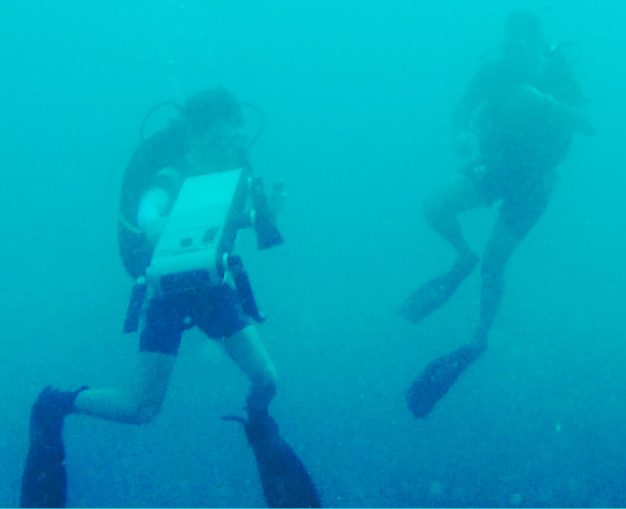}
    \caption{Two divers collaborating with an underwater robot. It is desirable for the robot to be able to uniquely identify its human partners; however, the divers' similar gear and the poor visibility conditions underwater make this difficult.}
    \label{fig:sim_divers}
\end{figure}

Most MOT research has focused on tracking pedestrians \cite{luo_multiple_2014, leal2017tracking, milan_mot16:_2016}. In this work, we adapt these pedestrian-focused strategies for use in an underwater human-robot collaboration scenario. In particular, this involves shifting the tracker's focus from tracking many people that come and go throughout a crowded scene, to tracking a small group of people that may leave the robot's field of view for an arbitrary period of time, but remain in the scene indefinitely. Also, human body postures are predominantly in a horizontal orientation during the diver tracking scenario, which is not the case for pedestrians. Our tracking problem has some difficulties that are not present in the typical pedestrian tracking scenarios: there is the inherent difficulty in detecting divers (see Section~\ref{sec:related} for a discussion), as well as the difficulty in distinguishing between two divers given the poor visibility conditions underwater and similarities in divers' SCUBA gear (see Figure \ref{fig:sim_divers}). However, in our problem we only need to track a few divers at a time, whereas typical MOT sequences contain dozens of people in a given frame. 

Our approach is tracking-by-detection, which is the current leading MOT paradigm. In tracking-by-detection, the tracker first performs person detection on each image. The tracker aims to match each of these detections to the correct \textit{track}, where each track represents a unique person. Each track typically models the person's trajectory and/or appearance in order to assist the tracker in matching detections to tracks.

Our method is an extension of Simple Online Realtime Tracking with an appearance metric (deep SORT), a tracker that runs in realtime and performs well on standard tracking benchmarks. We extend this technique to use a custom diver appearance metric, allow tracks to persist even after arbitrarily long absences, and to recover from certain reidentification errors.

\begin{figure}
    \centering
    \includegraphics[trim={0 1cm 1cm 1cm},clip,width=0.9\linewidth]{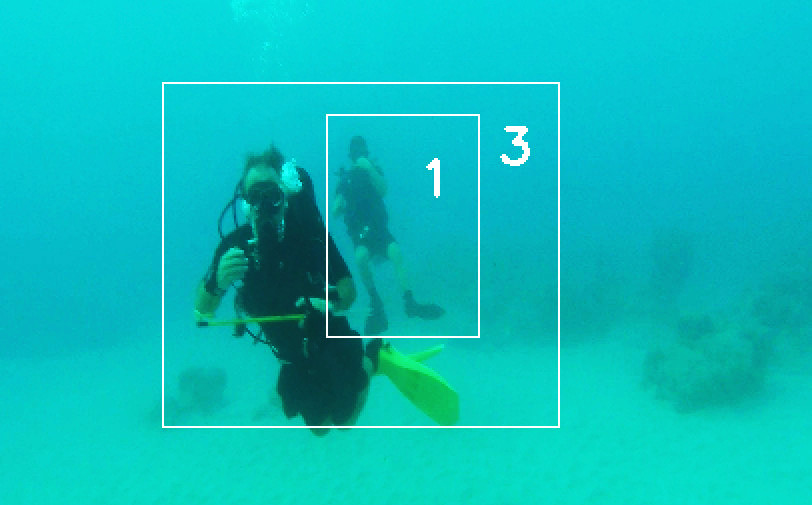}
    \caption{Exemplary output of our tracker on an underwater scene with multiple divers present. Two divers are detected and determined to belong to tracks ``1" and ``3."}
    \label{fig:example}
\end{figure}

Specifically, this paper contributes the following:
\begin{itemize}
    \item A multi-diver tracker that can run on a realtime robotic platform.
    \item An adaptation of the traditional multi-person tracking strategies, which focus on transient pedestrians in crowded scenes, to a strategy that focuses on tracking people who are always present in the scene (although they may not always be present in the robot's field of view).
    \item Evaluation of the proposed tracker on a diverse set of scenarios, including several that depict authentic underwater human-robot collaboration.
\end{itemize}

%% file: tex/related.tex
\section{Related Work}
\label{sec:related}
Distinguishing between different divers is typically a difficult task due to low in-class feature diversity: images of different divers are often highly visually similar, both because of similar wearables (\eg SCUBA gear) and poor visibility conditions underwater.  The authors' previous work includes a first-of-its-kind method to identify divers via k-means clustering on hand-crafted feature vectors \cite{xia2019visual}. The current research extends this work to an online method that can track and uniquely identify divers, utilizing a MOT approach.

MOT has been extensively studied in computer vision. Most research in the area has focused on tracking pedestrians \cite{milan_mot16:_2016, luo_multiple_2014, bernardin_research_2008, leal2017tracking}, and the annual MOT challenges primarily consist of pedestrian datasets \cite{milan_mot16:_2016, dendorfer2019cvpr19}. We refer the interested reader to \cite{luo_multiple_2014} for a thorough review of the field.

Many high-performing trackers process images in batches, rather than online, which makes them infeasible for realtime use~\cite{fagot2016improving, kim2015multiple, tang_multiple_2017}. Additionally, many of these trackers use computationally-intensive techniques such as optical flow analysis~\cite{xiang2015learning, shantaiya2015multiple} and Multiple Hypothesis Tracking (MHT) \cite{kim2015multiple} which increase tracker accuracy at the cost of processing speed. This phenomenon is illustrated by the leaderboard for the Conference on Computer Vision and Pattern Recognition (CVPR) 2019 MOT Challenge \cite{dendorfer2019cvpr19}. Only one of the three most accurate submissions has a tracking component that runs at more than two frames per second (the processing time for the detection component is not reported).

In contrast, the fastest trackers rely on relatively simple yet robust heuristics, while still achieving reasonable accuracy. Many of these realtime trackers do not take people's appearances into account, and instead match detections to tracks solely by analyzing the locations of the detections. For example, the Intersection Over Union (IOU) Tracker \cite{bochinski2017high} matches a detection to a given track if there is a sufficiently high IOU between the detection bounding box and the track's bounding box in the previous frame. Simple Online Realtime Tracking (SORT)  \cite{bewley2016simple} is a slightly more complex model that uses a Kalman filter \cite{kalman1960new} to model people's motions and predict their next location. SORT matches detections to tracks if there is a sufficiently high IOU between a detection's bounding box and the track's predicted bounding box. While these techniques are reasonable for tracking pedestrians, they are not able to reidentify a person after he or she is temporarily occluded. This shortcoming was addressed by deep SORT \cite{wojke_simple_2017}, which extends SORT to also use a deeply-learned appearance metric \cite{wojke_deep_2018} to improve reidentification after occlusions. This work extends deep SORT further to improve reidentification after longer occlusions or absences, as well as customizing the appearance metric for diver reidentification.

Reidentification in visual tracking is closely tied to the more general person reidentification (\ie reID) problem. The basic reID problem can be formulated as a task to compare persons of interest appearing in `query' datasets to a `gallery' of potential candidate images, captured from different angles, different cameras and even different scenes. Existing work focuses almost exclusively on person retrieval on land, which are either image-based or video-based. Person reID methods often use visual cues based on the individual's height, face, complexion, and gait~\cite{zajdel2005keeping,huang1997object}. However, these methods are not reliable in situations where face or gait recognition is not feasible (\eg for poor image resolution, or in images captured from different angles). Gheissari \textit{et al.} propose a novel method~\cite{gheissari2006person} which relies on features invariant to illumination, pose, and dynamic appearance of clothing~\cite{gheissari2006person}. Recent contributions increasingly rely on deep machine learning (\cite{li2014deepreid,xiao2017joint,zheng2017unlabeled,mclaughlin2016recurrent}) for their improved accuracy in the reID task, although most are not realtime capable.

Detection is another crucial component of tracking-by-detection systems. Diver detection is a difficult problem, largely because underwater visual perception presents various challenges, including color distortion, suspended particles, and light refraction, absorption, and scattering \cite{sattar2006performance, fabbri2018enhancing}. Additionally, divers have a wider range of potential positions and orientations than people on land, since divers are suspended in a 6-DOF aquatic environment. Islam \textit{et al.} \cite{islam2018toward} design a CNN-based realtime-capable diver detection model while sacrificing relatively little accuracy. We use this model for the detection component of our tracker.

%% file: tex/methodology.tex
\section{Methodology}
Our algorithm keeps a set of all known tracks, denoted as $\mathcal{T}$. After finding detections with the network described in \cite{islam2018toward} for a given frame, the algorithm attempts to match each detection to an existing track $t \in \mathcal{T}$. Below we describe how the algorithm makes these matches.

\subsection{Intersection Over Union}
Our algorithm first checks to see if any detections can be matched to tracks via intersection over union (IOU). We begin with this strategy because simple IOU trackers can be very effective \cite{bochinski2017high}, and IOU can be computed quickly. Therefore, we take the IOU ``shortcut" if possible before doing any more intensive computations.

We use the following definition for the IOU between bounding boxes $A$ and $B$:
$$
IOU(A, B) = \frac{A \cap B}{A \cup B}
$$

Our IOU-assignment procedure is as follows: We have a set of detections $\mathcal{D}_t$ that contains all detections found at time $t$. We then calculate the IOU between each $d_{i} \in \mathcal{D}_t$ and each $d_{j} \in \mathcal{D}_{t - 1}$. If $IOU(d_{i}, d_{j}) \geq 0.75$, $d_i$ is assigned to the same track as $d_{j}$. This is a more conservative threshold than the 0.5 threshold used by the IOU tracker \cite{bochinski2017high}, since the IOU tracker uses batch processing to eliminate erroneous associations and we process each frame sequentially.

\subsection{Appearance and Location Metrics}
To match the remaining detections to tracks, we utilize two metrics: an appearance metric that describes the similarity of a detection's appearance and a track's appearance, and a location metric that describes the similarity between a detection's location and a track's location.

\begin{figure}[t]
  \centering
  \begin{subfigure}[t]{0.4\linewidth}
    \includegraphics[trim={6cm 0.5cm 0 0},clip,width=\linewidth]{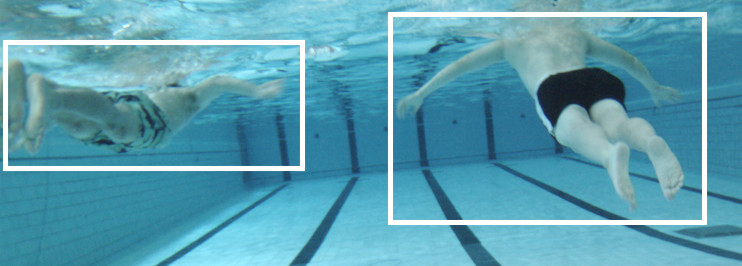}
  \end{subfigure}
  \begin{subfigure}[t]{0.4\linewidth}
    \includegraphics[trim={6cm 0.5cm 0 0},clip,width=\linewidth]{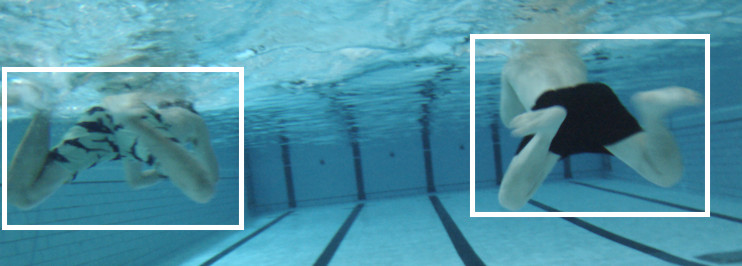}
  \end{subfigure}
  \caption{An illustration of how the area and aspect ratios of bounding boxes can oscillate rapidly during swimming.}
  \label{fig:bbox_demo}
\end{figure}

The \textbf{location metric} is largely identical to the approaches in \cite{wojke_deep_2018} and \cite{bewley2016simple}. We use a simple Kalman filter to model the motion of each track. Our tracking scenario is defined on the four-dimensional state space $(x, y, \dot{x}, \dot{y})$ where $(x, y)$ is the bounding box's center position in image coordinates and $\dot{x}$ and $\dot{y}$ are the respective velocities of $x$ and $y$. We do not include the area or aspect ratio of the bounding box in the state space because due to rapid arm and leg motion from swimming strokes, the aspect ratio and area of a diver's bounding box can oscillate rapidly (see Figure \ref{fig:bbox_demo} for an illustration of this effect). The location metric between detection $d_i$ and track $t_j$ is then defined as the squared Mahalanobis distance between the Kalman filter's predicted location of $t_j$ and the actual location of $d_i$.

The \textbf{appearance metric} differs from the approach used by other trackers. We calculate a series of hand-crafted features to find a feature vector that describes the appearance of the detected diver. In contrast, most trackers use a deep neural network trained on a person reidentification dataset to generate feature vectors, \eg \cite{su2016deep, wojke_deep_2018, long2018real}. We do not take this approach for two reasons: (a) networks trained on person reidentification datasets are not well-suited to reidentifying divers \cite{youya_reid}, and (b) diver-specific reidentification datasets do not exist and data scarcity prevents us from creating one.

The appearance features we extract and the reasoning behind their inclusion are fully described in~\cite{xia2019visual}. In summary, we use features that can satisfactorily differentiate between divers, but are also relatively robust to changes in lighting, and diver position and orientation:
\begin{itemize}
    \item Average color distribution in the LAB color space
    \item Amplitude of the spatial frequency distribution
    \item Shape approximation through image contours
    \item Shape approximation through convex hull
    \item Hu image moment invariants \cite{hu1962visual}
\end{itemize}
These features have been shown to be sufficient for the k-means algorithm to effectively cluster images of divers according to their identities \cite{xia2019visual}.

\begin{figure*}[hbt]
  \centering
  \begin{subfigure}[t]{0.32\linewidth}
    \includegraphics[width=\linewidth]{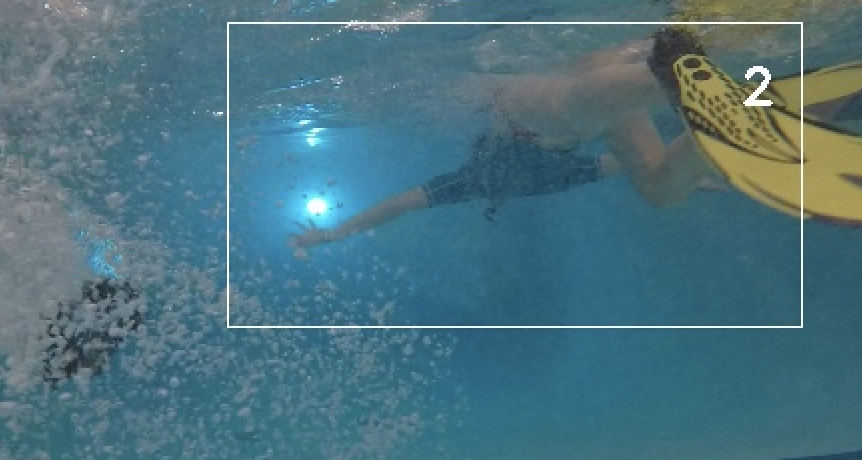}
     \caption{}
     \vspace{4mm}
     \label{fig:id_rec_a}
  \end{subfigure}
  \begin{subfigure}[t]{0.32\linewidth}
    \includegraphics[width=\linewidth]{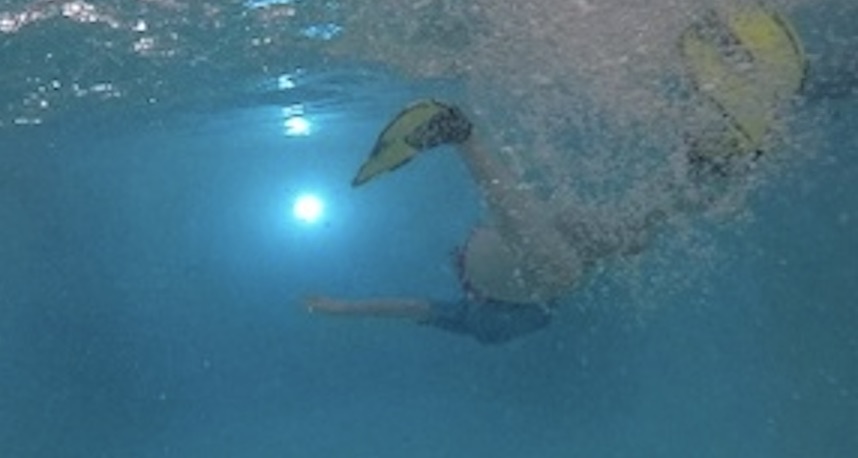}
    \caption{}
    \label{fig:id_rec_b}
    \vspace{4mm}
  \end{subfigure}
  \begin{subfigure}[t]{0.32\linewidth}
    \includegraphics[width=\linewidth]{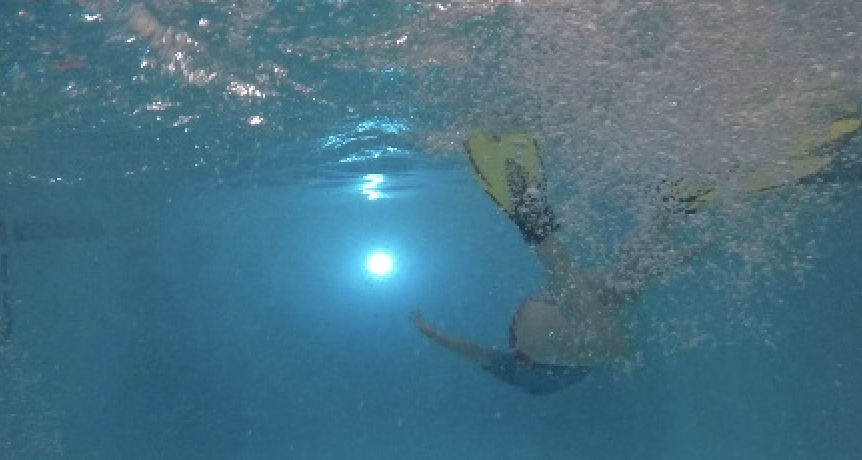}
    \caption{}
    \label{fig:id_rec_c}
    \vspace{4mm}
  \end{subfigure}
  \begin{subfigure}[t]{0.32\linewidth}
    \includegraphics[width=\linewidth]{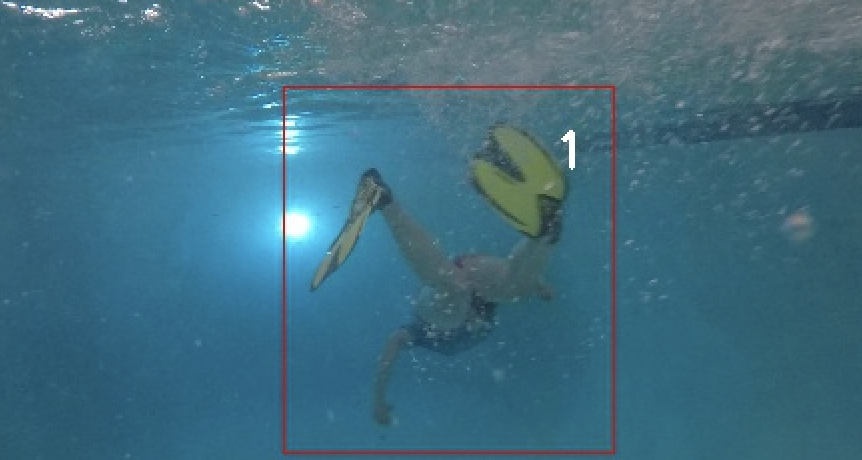}
    \caption{}
    \label{fig:id_rec_d}
  \end{subfigure}
  \begin{subfigure}[t]{0.32\linewidth}
     \includegraphics[width=\linewidth]{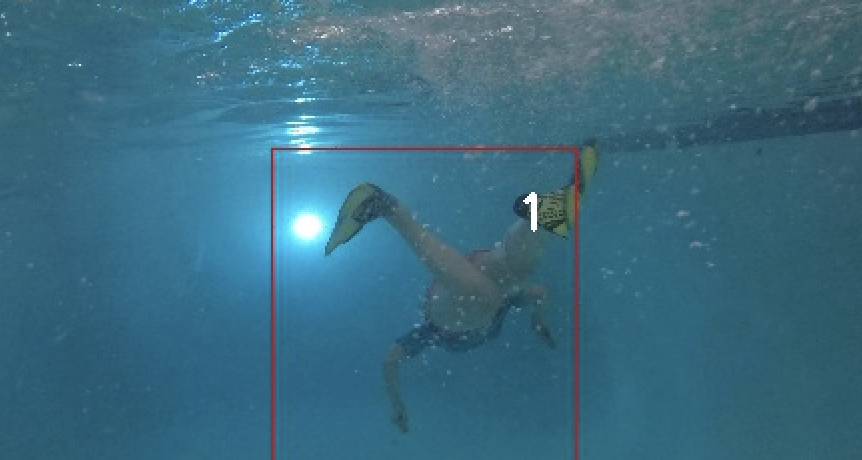}
     \caption{}
     \label{fig:id_rec_e}
  \end{subfigure}
    \begin{subfigure}[t]{0.32\linewidth}
     \includegraphics[width=\linewidth]{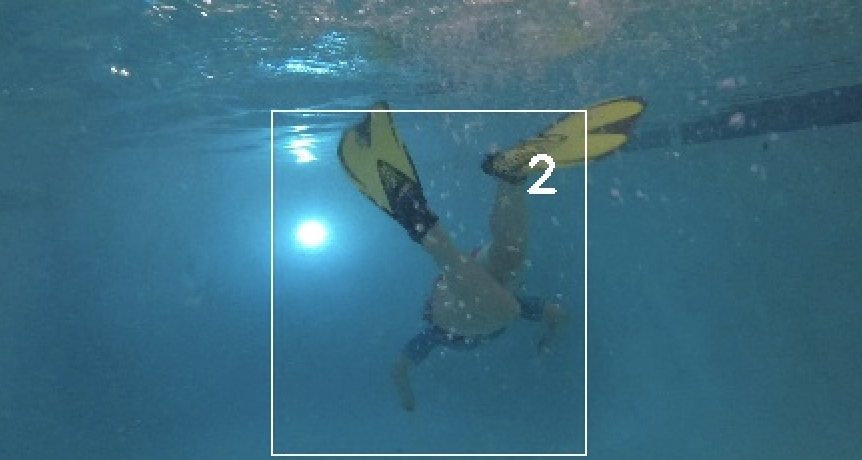}
     \caption{}
     \label{fig:id_rec_f}
  \end{subfigure}
  \caption{An example of our identity recovery technique. (\ref{fig:id_rec_a}) This swimmer is being tracked with identity ``2.'' (\ref{fig:id_rec_b}-\ref{fig:id_rec_c}) When the swimmer descends, bubbles partially occlude the body and the detector fails. (\ref{fig:id_rec_d}) Once the bubbles begin to clear, the change in body position means that the tracker does not find enough appearance similarity to correctly match this detection to track ``2." Instead a new track ``1" is created. (\ref{fig:id_rec_e}) Since ``1" is a new track, the algorithm continually checks for sufficient similarities between ``1" and tracks that have not been seen since ``1" appeared. (The red colored bounding box indicated that the tracker is uncertain about this track's identity.) (\ref{fig:id_rec_f}) The algorithm finds a sufficient amount of similarity between new track ``1" and old track ``2." Track ``1" is merged into ``2", and the swimmer's appropriate identity is recovered.}
  \label{fig:id_rec}
\end{figure*}

We take the same approach as \cite{wojke_simple_2017} for calculating appearance similarities between tracks and detections. For each track, we store the normalized feature vectors of the $100$ most recent detections that have been matched to that track. To measure the appearance similarity between a detection and a track, we find the cosine similarities between each of the track's stored feature vectors and that detection's normalized feature vector. The smallest of these cosine similarities is then the appearance similarity between the track and detection. Concretely, if detection $d_i$'s normalized feature vector is $\bm{f_i}$, and $\mathcal{F}_j$ is the set of stored normalized feature vectors for track $t_j$, then the appearance similarity between $d_i$ and $t_j$ is calculated with: $$
\text{sim}_\text{appearance}(d_i, t_j) = \min(1 - \bm{f_{i}}^T\bm{f_j} \hspace{3pt} | \hspace{3pt} \bm{f_j} \in \mathcal{F}_j)
$$

\subsection{Matching Detections to Tracks}
Next, we must use the location and appearance metrics to match the remaining detections to tracks.
We use the typical strategy of formulating an assignment problem that can be solved with the Hungarian algorithm \cite{kuhn1955hungarian}. This is done by finding a cost, $c_{ij}$ for matching detection $d_i$ detection to track $t_j$. We let $c_{ij} = \text{sim}_\text{appearance}(d_i, t_j)$. Because of poor visibility underwater and strong resemblance between divers' SCUBA gear, sometimes a detection is highly visually similar to more than one track. In this case, we introduce the location similarity as a tie-breaker and have $c_{ij} = \text{sim}_\text{appearance}(d_i, t_j) + \text{sim}_\text{location}(d_i, t_j)$ for all costs associated with that detection. The Hungarian algorithm then finds the optimal matches between detections and tracks such that the costs are minimized.

We do not rely heavily on the location similarity to contribute to the cost of a match. This is because the location similarity is derived through a Kalman filter, which is designed to model linear systems \cite{kalman1960new}. Since divers' movements are not consistently linear, we cannot highly depend on Kalman filter predictions. In addition, the robot's exact motions (and by extension the camera's exact motions) are unknown, which also negatively affects the Kalman filter's predictive power.

However, the Kalman filter can be useful for determining which matches between detections and tracks are unacceptable. We consider matches between detections and tracks to be unacceptable if the detection's location is too far away from the track's predicted location (\ie if the location metric is above a certain threshold), or if the detection's appearance is too dissimilar from the track's appearance (\ie if the appearance metric is above a certain threshold). The thresholds used for unacceptable matches were found empirically by testing our algorithm on a validation dataset. We used $25$ for the location metric threshold and $1$e$-4$ for the appearance metric threshold. 

We indicate an unacceptable match by setting $c_{ij} = \infty$. If a detection cannot be matched to a track with a cost $c < \infty$, we create a new track for that detection. To account for spurious detections, new tracks are not officially included in $\mathcal{T}$ until they have been matched with a detection for three consecutive frames.

\subsection{Short Term vs. Long Term Reidentification}
When a detection is not matched to a track for a frame, the track is no longer active. If a person belonging to an inactive track is detected, the person will need to be reidentified, \ie matched to their existing track.

Our algorithm as described above can accomplish reidentification if the detection's appearance is similar enough to the inactive track and its location is similar enough to the inactive track's predicted location. However, our algorithm changes its approach for tracks that have been inactive for a longer period of time (\ie more than five frames). In this case, we no longer calculate a predicted track location, because the track's Kalman filter will have too much uncertainty and propagated error. We also increase the appearance similarity threshold slightly, to $5$e$-4$. This is because when a person is absent from the scene for a longer period of time, there may be significant changes to their position and orientation, as well as the scene's lighting, so we adopt a more forgiving threshold. The threshold was also obtained empirically by testing our algorithm on a validation dataset.

\subsection{Identity Recovery}
One potential problem with our reidentification technique is that we must correctly reidentify a diver on the first frame in which he or she reenters the robot's field of view. If, in that initial frame, the diver's position or a temporary partial occlusion leads to a high dissimilarity between the detection's appearance metric and the diver's true track, the diver will not be assigned to their true track. We refer to this scenario as a missed reidentification.

Missed reidentifications are not generally a large concern in the MOT community, and the standard MOT evaluation metrics do not heavily penalize missed reidentifications. However, for our use case, missed reidentifications are hugely problematic since they lead to the robot being mistaken about a diver's identity. To address this problem, we have a procedure to correct missed reidentifications.

The process is as follows: after the algorithm matches detections to tracks, it examines the set of new tracks. We consider a track to be ``new" if it has existed for 15 or fewer frames. For each new track $t_i$, the algorithm finds the set of tracks that have not been seen since $t_i$ was created. We consider this set to contain all tracks that could possibly belong to the same person represented by $t_i$. 

Next, the algorithm checks to see if $t_i$ does in fact represent the same person as another track $t_j$. To do this, we find the cosine similarity between each feature vector stored for $t_i$ and each feature vector stored for $t_j$. If more than 25\% of these cosine similarities are below our acceptable appearance similarity matching threshold, the algorithm decides that $t_i$ and $t_j$ share an identity. (Again, this threshold was established empirically through testing on a validation dataset.) Then track $t_i$ is merged into $t_j$. See Figure \ref{fig:id_rec} for an illustration of this technique in practice.

%% file: tex/experiments.tex
\section{Experiments}

We evaluate the performance of the proposed tracker by testing it on eight videos, six of which take place in open-water (\ie ocean) environments, and two of which take place in closed-water (\ie pool) environments (see Table~\ref{table:results}). The six ocean videos were all recorded during field trials and thus represent authentic underwater human-robot collaboration scenarios. The two pool videos were recorded to simulate human-robot collaboration scenarios. Each video contains between two and four divers, and the videos are about 10 minutes long in total. The ground truth identities and bounding boxes for each video were annotated by hand. 

\begin{figure}[bth]
	\centering
	\includegraphics[width=\linewidth]{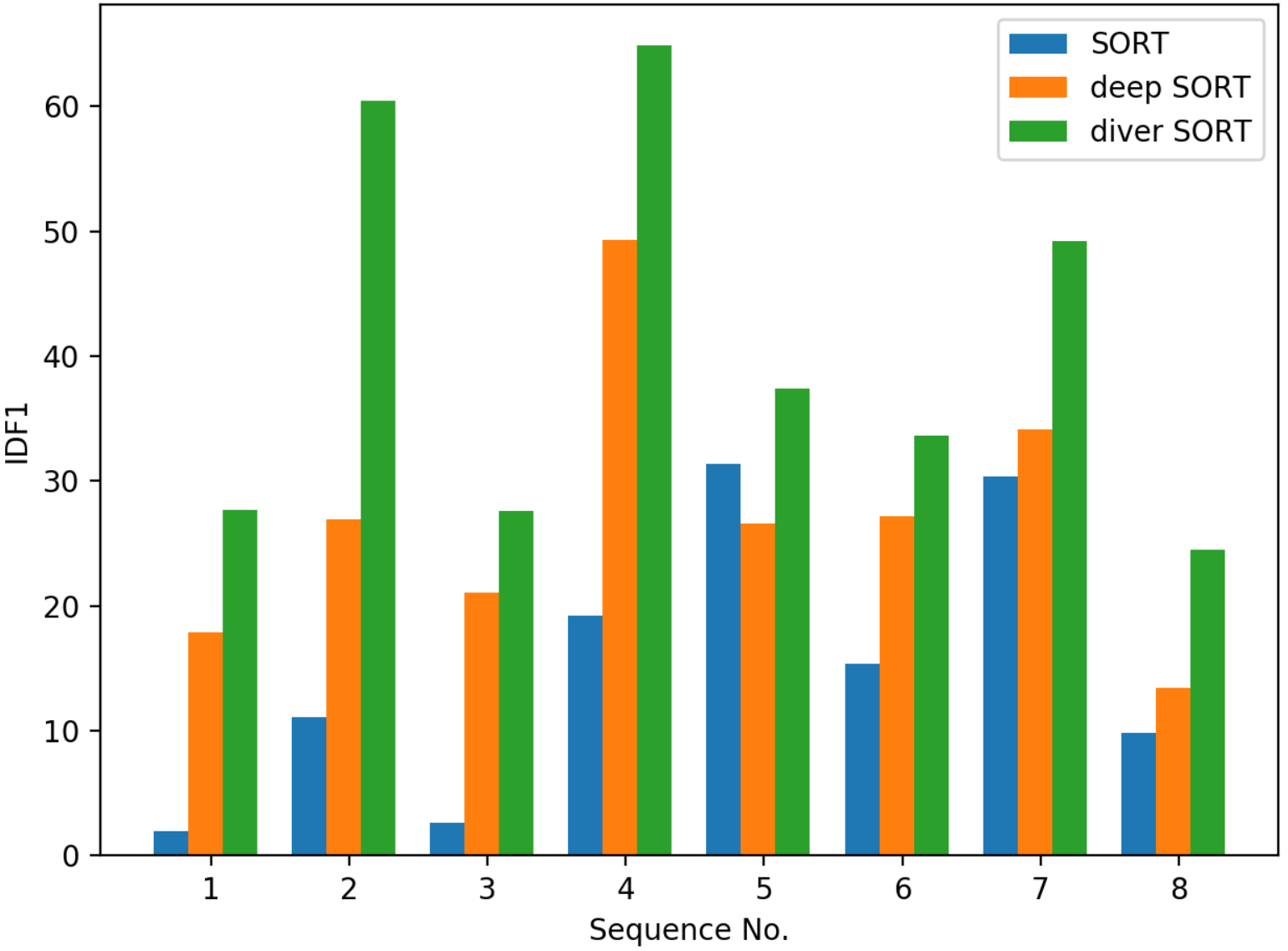}
	\caption{A comparison of the IDF1 metric across all eight scenarios for the three trackers tested.}
	\label{fig:bar_graph}
\end{figure}

 We compare our tracker's performance to SORT and deep SORT to ensure that our modifications result in better tracking for underwater human-robot collaboration scenarios. The SORT method does not incorporate any appearance information, whereas deep SORT uses a CNN trained on the MARS person reidentification dataset \cite{zheng2016mars}. Table~\ref{table:results} contains a summary of the three trackers' performances on our eight videos. All trackers used our custom diver detector to generate detections (note that the detector's performance is also included Table~\ref{table:results}). We then used standard MOT metrics \cite{ristani2016MTMC} to evaluate how well the different trackers identify divers; see Table~\ref{table:metrics} for a brief description of the metrics.
 
\begin{table}[!hbt]
    {\renewcommand{\arraystretch}{1.5}
	\begin{tabular}{@{}l |l @{}p{0.8\linewidth}}
		\textbf{Metric} & & \textbf{Description} \\
		\hline
		DP   & & Detection Precision. Defined as $\text{TP}/(\text{TP} + \text{FP})$ where TP is a true positive (\ie a detection that closely matches a ground truth bounding box) and FP is a false positive (\ie a detection that does not closely match a ground truth bounding box). All trackers used the same detector.  \\
		\hline 
		DR   & & Detection Recall. Defined as $\text{TP}/(\text{TP} + \text{FN})$ where TP is a true positive detection and FN is a false negative detection (\ie a ground truth bounding box that does not closely match any detection). All trackers used the same detector \\
		\hline 
		IDF1 & & Identity F1. Harmonic mean of IDR and IDP. \\
		\hline 
		IDP  & & Identity Precision. Defined as $(\text{IDTP})/(\text{IDTP} + \text{IDFP})$ where IDTP is the number of true positive identities (\ie identities output by the tracker that match ground truth identities) and IDFP is the number of false positive identities (\ie identities output by the tracker that do not match ground truth identities) in the tracker's output. \\
		\hline 
		IDR  & & Identity Recall. Defined as $(\text{IDTP})/(\text{IDTP} + \text{IDFN})$ where IDTP is the number of true positive identities, IDFP is the number of false positive identities, and IDFN is the number of false negative identities. \\
		\hline 
		IDS  & & Identity Switches. The total of number of times that a tracked trajectory changes its matched ground truth identity. \\
		\hline 
		FM   & & Fragmentations. The total number of times a tracked trajectory is interrupted (\ie frames are dropped).  
	\end{tabular}}
	\caption{Description of MOT metrics used in evaluation. See \cite{ristani2016MTMC, leal2015motchallenge} for more detailed discussions.}
	\label{table:metrics}
\end{table}

\begin{table*}[!htb]
\centering
{\renewcommand{\arraystretch}{1}
\begin{tabular}{ cclcc || lccccc}
    \multicolumn{1}{p{.31in}}{Scen.} & \multicolumn{1}{c}{Location} & \multicolumn{1}{p{2.2cm}}{Exemplar} & \multicolumn{1}{c}{DP $\uparrow$} & \multicolumn{1}{c}{DR $\uparrow$} 
        & {Tracker} & {IDF1 $\uparrow$} & {IDP $\uparrow$} & {IDR $\uparrow$} & {IDS $\downarrow$} & {FM $\downarrow$} \\
    \midrule \midrule

    \multirow{5}{*}{\textbf{1}} & \multirow{5}{*}{\textbf{ocean}} & \multirow{5}{*}{\parbox[c]{1.25em}{
    		\includegraphics[width=5.75\linewidth]{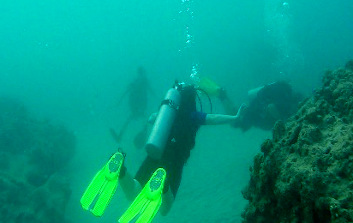}}} & \multirow{5}{*}{49.7} & \multirow{5}{*}{38.7} & & & & & & \\
    		
    & & & & & SORT & 2.0 & 28.6 & 1.0 & \textbf{0} & \textbf{0} \\
     & & & & & deep SORT & 17.9 & 20.5 & 15.8 & 22 & 25 \\
	& & & & & diver SORT & \textbf{27.7} & \textbf{39.3} & \textbf{21.4} & 7 & 19 \\
	& & & & & & & & & & \\
    \bottomrule
    \multirow{5}{*}{\textbf{2}} & \multirow{5}{*}{\textbf{ocean}} & \multirow{5}{*}{\parbox[c]{1.25em}{
    		\includegraphics[width=5.75\linewidth]{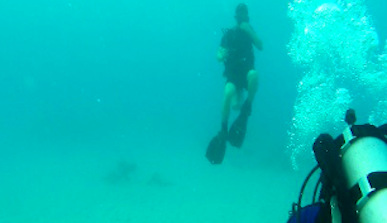}}} & \multirow{5}{*}{59} & \multirow{5}{*}{35.9} & & & & & & \\
    		
    & & & & & SORT & 11.1 & 57.1 & 6.2 & 1 & \textbf{1} \\
	& & & & & deep SORT & 26.9 & 35.9 & 21.5 & 3	& 7 \\
	& & & & & diver SORT & \textbf{60.4} & \textbf{78.0} & \textbf{49.2} & \textbf{0} & 8 \\
	& & & & & & & & & & \\
    \bottomrule
    
    \multirow{5}{*}{\textbf{3}} & \multirow{5}{*}{\textbf{ocean}} & \multirow{5}{*}{\parbox[c]{1.25em}{
    		\includegraphics[width=5.75\linewidth]{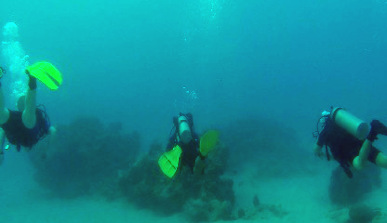}}} & \multirow{5}{*}{64.5} & \multirow{5}{*}{49.1} & & & & & & \\
    & & & & & SORT & 2.6 & \textbf{42.9} & 1.3 & \textbf{0} & \textbf{0} \\
	& & & & & deep SORT & 21.1 & 24.4 & 18.5 & 24 & 29 \\
	& & & & & diver SORT & \textbf{27.6} & 41.6 & \textbf{20.7} & 15 & 27 \\
	& & & & & & & & & & \\
    \bottomrule
    
    \multirow{5}{*}{\textbf{4}} & \multirow{5}{*}{\textbf{ocean}} & \multirow{5}{*}{\parbox[c]{1.25em}{
    		\includegraphics[width=5.75\linewidth]{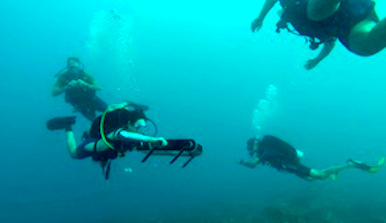}}} & \multirow{5}{*}{62.7} & \multirow{5}{*}{37.2} & & & & & & \\
    & & & & & SORT & 15.4 & 25.3 & 11.1 & \textbf{11} & \textbf{25} \\
	& & & & & deep SORT & 27.2 & 36.6 & 21.7 & 16 & 26 \\
	& & & & & diver SORT & \textbf{33.6} & \textbf{46} & \textbf{26.5} & 19 & 46 \\
	& & & & & & & & & & \\
    \bottomrule
    
    \multirow{5}{*}{\textbf{5}} & \multirow{5}{*}{\textbf{ocean}} & \multirow{5}{*}{\parbox[c]{1.25em}{
    		\includegraphics[width=5.75\linewidth]{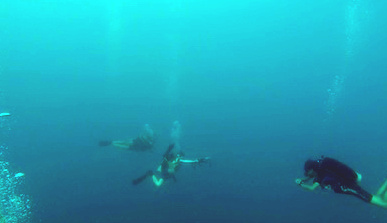}}}  & \multirow{5}{*}{70.7} & \multirow{5}{*}{54.7} & & & & & & \\
    & & & & & SORT & 30.4 & 42.2 & 23.8 & 23 & \textbf{63} \\
	& & & & & deep SORT & 34.1 & 39.2 & 30.3 & 30 & 80 \\
	& & & & & diver SORT & \textbf{49.2} & \textbf{58.6} & \textbf{42.4} & \textbf{19} & 92 \\
	& & & & & & & & & & \\
    \bottomrule
    
   \multirow{5}{*}{\textbf{6}} & \multirow{5}{*}{\textbf{ocean}} & \multirow{5}{*}{\parbox[c]{1.25em}{
   		\includegraphics[width=5.75\linewidth]{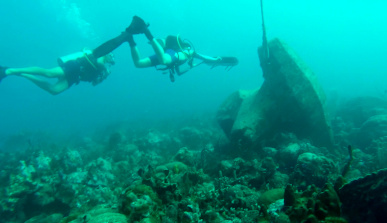}}} & \multirow{3}{*}{30.5} & \multirow{3}{*}{25}  & & & & & & \\
   	& & & & & SORT & 9.8 & 12.9 & 7.9 & 15 & \textbf{37} \\
	& & & & & deep SORT & 13.4 & 15.4 & 11.9 & \textbf{9} & \textbf{37} \\
	& & & & & diver SORT & \textbf{24.5} & \textbf{28.4} & \textbf{21.5} & 10 & 53 \\
	& & & & & & & & & & \\
    \bottomrule
    
    \multirow{5}{*}{\textbf{7}} & \multirow{5}{*}{\textbf{pool}} & \multirow{5}{*}{\parbox[c]{1.25em}{
    		\includegraphics[width=5.75\linewidth]{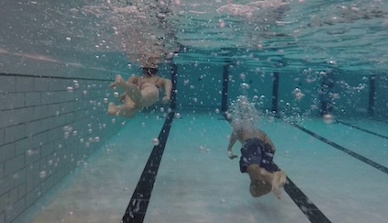}}} & \multirow{3}{*}{81.6} & \multirow{3}{*}{60.3} & & & & & & \\
    & & & & & SORT & 19.2 & 27.7 & 14.7 & 25 & \textbf{43}\\
	& & & & & deep SORT & 49.3 & 58.1 & 42.8 & \textbf{14} & 53 \\
	& & & & & diver SORT & \textbf{64.9} & \textbf{80.0} & \textbf{54.6} & 17 & 65 \\
	& & & & & & & & & & \\
    \bottomrule
    
    \multirow{5}{*}{\textbf{8}} & \multirow{5}{*}{\textbf{pool}} & \multirow{5}{*}{\parbox[c]{1.25em}{
    		\includegraphics[width=5.75\linewidth]{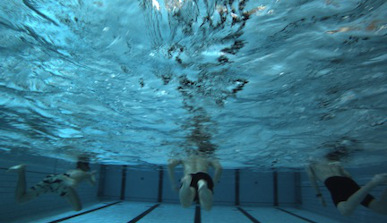}}}  & \multirow{3}{*}{63.2} & \multirow{3}{*}{25.8} & & & & & & \\
    & & & & & SORT & 31.4 & \textbf{68.6} & 20.3 & \textbf{2} & \textbf{13} \\
	& & & & & deep SORT & 26.6 & 46.3 & 18.6 & 4 & 15 \\
	& & & & & diver SORT & \textbf{37.4} & 51.9 & \textbf{29.2} & 12 & 26 \\
	& & & & & & & & & & \\
    \midrule\midrule
    \multirow{3}{*}{\textbf{ALL}} & & & & & SORT & 18.6 & 29.6 & 13.6 & \textbf{77} & \textbf{182} \\
    & & & & & deep SORT & 30.0 & 36.3 & 25.5 & 122 & 272 \\
    & & & & & diver SORT & \textbf{42.1} & \textbf{53.1} & \textbf{34.9} & 99 & 336 \\
    \bottomrule
\end{tabular}}
    \caption{A comparison of our algorithm's performance (diver SORT) and two other realtime trackers' performance on several videos of divers. See Table~\ref{table:metrics} for a brief description of the metrics used. }
    \label{table:results}
\end{table*}

Across all eight videos, our tracker performed best in correctly identifying detected divers as measured by IDF1 (Figure~\ref{fig:bar_graph}), IDP, and IDR. This indicates that our reidentification and identification recovery methods are effective. Since our tracker specifically outperforms deep SORT, the results also indicate that deeply learned person reidentification appearance metrics fall short of the hand-crafted features used by our tracker. However, deep SORT does consistently outperform SORT on identification, so the deeply learned appearance metric is not ineffective.

Our tracker had no improvement over the others on identity switches and fragmentations: at best, our tracker had marginally fewer identity switches (\eg sequences 2 and 5 in Table~\ref{table:results}); otherwise, it was not a top performer. Our tracker also consistently had the most fragmentations, likely because our tracker persists track identities even through long occlusions or periods of absence. In some cases, such as Scenario 1, the baseline SORT tracker had extremely low IDS and FM scores. This was due to SORT only matching a few detections to tracks, and each track lasting for only 1-2 frames. Such tracks are too short to experience identity switches and fragmentations.

It is also important to note that there was a wide range for IDF1, IDP, and IDR values across the eight videos. Additionally, there is a strong relationship between those values and the detector's performance. For example, video 7 resulted in the best detector performance, as well as our tracker's best IDF1 score ($64.9$). On the other hand, video 6 resulted in our detector's worst performance and also our tracker's worst IDF1 score ($24.5$). We found a positive correlation between detection precision and identity precision for both our tracker (Pearson $r \approx 0.727, p < 0.05$) and deep SORT (Pearson $r \approx 0.850, p < 0.05$). This is unsurprising since detection quality is known to have a high impact on a tracker's performance \cite{luo_multiple_2014, bewley2016simple, leal2015motchallenge}.

Our tracker's end-to-end image processing rate is 9.8 frames per second on a machine with an AMD Ryzen 5 2600 3.9GHz processor and 16GB of RAM, making it realtime capable for AUV deployment.

%% file: tex/conclusion.tex
\section{Conclusion}
In this paper, we propose a method to allow underwater robots to track and identify people in real time. Our method combines techniques from multi-object tracking and underwater diver detection and identification. Results show that our custom tracker has more correct identifications than baseline realtime trackers on underwater datasets. However, in situations where the detector produces highly inaccurate detections, all tested trackers perform poorly. This has two implications: (1) improving our detector can result in highly significant tracking improvements, and (2) the tracker has the potential to be highly unreliable in the field when adverse conditions reduce detection accuracy. In such situations, the erroneous tracker output could lead to erratic and unpredictable robot behavior, which may jeopardize the mission. Future work therefore involves improving realtime diver detection and developing a system that can flag poor detector performance (\eg dropping many frames, inconsistent numbers of detections between frames). When poor detection conditions are identified, the humans working with the robot can be made aware that the robot's tracking module should not be relied upon.